\documentclass[conference]{IEEEtran}
\IEEEoverridecommandlockouts

\usepackage{import}
\usepackage{subcaption}
\usepackage{graphicx}
\usepackage{threeparttable}
\usepackage{amsmath}
\usepackage{todonotes}
\usepackage{balance}
\usepackage{siunitx}

\def\BibTeX{{\rm B\kern-.05em{\sc i\kern-.025em b}\kern-.08em
    T\kern-.1667em\lower.7ex\hbox{E}\kern-.125emX}}
\begin{document}






\title{Towards Real-World Deployment of\\ Reinforcement Learning for Traffic Signal Control
\thanks{This work is part of the KI4LSA project and was supported by the German Federal Ministry of Transport and Digital Infrastructure (BMVI).}
}

\author{\IEEEauthorblockN{Arthur Müller, Vishal Rangras, Tobias Ferfers, Florian Hufen, Lukas Schreckenberg, Jürgen Jasperneite}
\IEEEauthorblockA{
\textit{Fraunhofer IOSB-INA}\\
Lemgo, Germany \\
\{givenname.surname\}@iosb-ina.fraunhofer.de}
\and
\IEEEauthorblockN{Georg Schnittker, Michael Waldmann}
\IEEEauthorblockA{
\textit{Stührenberg GmbH}\\
Detmold, Germany \\
\{g.schnittker,m.waldmann\}@stuehrenberg.de}
\and
\IEEEauthorblockN{Maxim Friesen}
\IEEEauthorblockA{
\textit{OWL University of Applied Sciences and Arts}\\
Lemgo, Germany \\
maxim.friesen@th-owl.de}
\and
\IEEEauthorblockN{Marco Wiering}
\IEEEauthorblockA{
\textit{University of Groningen}\\
Groningen, The Netherlands \\
m.a.wiering@rug.nl}
}

\maketitle

\begin{abstract} 
Sub-optimal control policies in intersection traffic signal controllers (TSC) contribute to congestion and lead to negative effects on human health and the environment. Reinforcement learning (RL) for traffic signal control is a promising approach to design better control policies and has attracted considerable research interest in recent years.
However, most work done in this area used simplified simulation environments of traffic scenarios to train RL-based TSC. To deploy RL in real-world traffic systems, the gap between simplified simulation environments and real-world applications has to be closed.
Therefore, we propose LemgoRL, a benchmark tool to train RL agents as TSC in a realistic simulation environment of Lemgo, a medium-sized town in Germany. In addition to the realistic simulation model, LemgoRL encompasses a traffic signal logic unit that ensures compliance with all regulatory and safety requirements. LemgoRL offers the same interface as the well-known OpenAI gym toolkit to enable easy deployment in existing research work. 
To demonstrate the functionality and applicability of LemgoRL, we train a state-of-the-art Deep RL algorithm on a CPU cluster utilizing a framework for distributed and parallel RL and compare its performance with other methods. 
Our benchmark tool drives the development of RL algorithms towards real-world applications. 



\end{abstract}
\begin{IEEEkeywords}
deep reinforcement learning, traffic signal control, intelligent transportation system, traffic simulation
\end{IEEEkeywords}

\section{Introduction}
Urban traffic is mainly controlled by intersection \textit{traffic signal controllers} (TSC). Most TSC are actuator-based, i.e. they choose the traffic light signals according to the current traffic demand. Conventional TSC policies in most cases represent a set of many rules implemented by traffic engineers. Due to the dynamic and stochastic nature of traffic, manually designed TSC policies cannot control traffic optimally in all situations. This leads to inefficient use of existing infrastructure and thus to congestion, which has negative effects on the environment and human health \cite{Pandian2009}. 
It is also a huge economic problem. According to \cite{Schrank2019}, the cost of traffic congestion in 2017 in urban areas of the United States is estimated at \$179 billion.             

\textit{Reinforcement Learning} (RL), a long-standing research area in the field of artificial intelligence, is a promising approach to overcome the limitations of manually designed TSC policies. 
In RL an agent acting with an environment tries to optimize its behavior in order to achieve an optimization goal (e.g. smooth traffic flow). RL for traffic signal control has been extensively researched over the last two decades. In particular, the combination of deep learning and RL, referred to as Deep RL, has initiated an increased interest in RL for traffic signal control. This is because Deep RL has led to breakthroughs in solving complex problems such as the game of Go \cite{Silver2017}.
As a result of the extensive research made, there are several RL-based approaches in literature that have been shown to be superior to conventional TSC in simulation environments \cite{Yau2017, Haydari2020}.
Despite all efforts in this field, there is no real-world deployment of an actual RL-based TSC to date. In fact, it has not yet been proven that RL is even applicable as TSC in a real-world setting. 
Every investigation so far has been done only in simplified simulation environments~\cite{Haydari2020}. To close the gap from training an RL agent in a simplified environment to deploy an RL-based TSC in the real-world, a realistic simulation environment is required as an intermediate step.  



To this end, we introduce \textit{LemgoRL}, a benchmark  tool  to  train  RL agents for traffic signal control in a realistic simulation scenario of an intersection in Lemgo, a medium-sized town in Germany. LemgoRL provides: 
\begin{itemize}
    \item a realistic model of a road network in Lemgo,
    \item simulation of all road user groups (vehicles, cyclists, and pedestrians) and not only considering vehicles,
    \item simulation of different vehicle classes (cars, motorcyclists, trucks, trucks with trailers, buses) and not only considering cars,
    \item simulation of the traffic volume according to measured data and not synthetic data,
    \item a set of realistic traffic phases designed by traffic engineers (domain experts),
    \item a mechanism to carry out transitions from one traffic light phase to another in a realistic manner, taking into account safety and regulatory requirements,
    \item and a mechanism to ensure compliance with all safety and regulatory requirements (e.g. adherence to the minimum duration of traffic light phases) at all times during execution.
\end{itemize}
To the best of our knowledge, there is no simulation environment available in literature offering this degree of realism and complexity for RL-based TSC. LemgoRL is based on the programming language Python and offers the same interface like the well-known OpenAI Gym toolkit~\cite{Brockman2016}, allowing RL researchers easy integration into their work. It is an intermediate step in our research work towards the deployment of an RL-based TSC at a real-world intersection in the city of Lemgo. We provide LemgoRL as an open-source tool at https://github.com/rl-ina/lemgorl.

\section{Preliminaries}
\subsection{Reinforcement Learning}
In RL an \emph{agent} is interacting with its \emph{environment}.  
At each time step $t$, the agent gets information about the current situation of the environment called \emph{state} $s_t$. Based on this information the agent selects the next \emph{action} $a_t$ to take in the environment. The behavior or decisions made by the agent are determined by the current \emph{policy} $\pi$. It is a function that maps states to actions $\pi(a_t | s_t)$. Additionally, at each time step the agent receives a \emph{reward} signal $r_{t+1}$ from the environment, which is a scalar value. The goal of the agent is to maximize the cumulative reward in the long run. 


Typically RL problems are formalized as \textit{Markov decision processes} (MDPs). An MDP is a tuple consisting of the following elements \cite{Sutton2018}:
\begin{itemize}
    \item A set of states $\mathcal{S}$,
    \item a set of actions $\mathcal{A}$,
    \item a state transition probability function $\mathcal{T}(s_t, a_t, s_{t+1})$ which defines a probability distribution over the successor state $s_{t+1}$ given the former state $s_t$ and action $a_t$,  
    \item a reward function $\mathcal{R}(s_t,a_t)$ determining the reward $r_{t+1}$, when taking action $a_t$ in state $s_t$,  
    \item and a discount factor $\gamma \in [0, 1]$, discounting future rewards.
\end{itemize}
A central requirement for an MDP is the so-called Markov property, which assumes that the current state $s_t$ contains all relevant
information of the past states and actions. Thus, the successor state $s_{t+1}$ only depends on the current state and action
and not on the past states and actions. 
Furthermore, MDP models can represent \emph{continuous}, meaning endless tasks, or \emph{episodic} tasks in which there is a terminal state.

\subsection{Traffic Simulator}
In order to train RL agents for TSC, a simulation environment is needed.
This is because, sampling data in reality would lead to sub-optimal decisions during training and therefore unacceptable waiting times for road users. 
RL algorithms are usually sample-inefficient, which would make training too slow. There are several traffic simulation tools that can provide the required simulation environment. According to \cite{Krauss1998}, they can be categorized as macroscopic, mesoscopic, microscopic and submicroscopic simulators, with increasing levels of detail. For training and testing of TSC, usually microscopic traffic simulators are chosen which provide a reasonable tradeoff between simulation speed and required level of detail. Among these, there are commercial software tools like Vissim \cite{Fellendorf1994}, Aimsun \cite{Casas2010}, and Quadstone Paramics \cite{Cameron1996} and free and open-source tools like MATSim \cite{HorniA.NagelK.2016} and SUMO \cite{Lopez2018}. 

For our simulation model, we use SUMO, mainly for two reasons.
First, according to a recent survey paper \cite{Haydari2020}, it is the most used tool for training RL as TSC. Over 70\% of all analyzed papers about Deep RL approaches for TSC used SUMO in their work. Since we want LemgoRL to be highly applicable and easy to use for the research community, choosing a well-known and widely used traffic simulator helps achieving this goal. Second, SUMO provides an extensive interface to Python called TraCI\footnote{https://sumo.dlr.de/docs/TraCI.html} (Traffic Control Interface). This allows retrieving and setting values for most simulated objects during simulation runtime (e.g. speed of a vehicle), which is necessary to extract the relevant information for the MDP-setting.

\section{Design of LemgoRL}
        
LemgoRL consists of several software components including a simulation model made with SUMO, a traffic signal logic unit (TSLU) implemented with the traffic engineering software LISA, a middleware, and an environment controller, both implemented in Python. The environment controller encapsulates the components as a single Python class and provides an interface to the agent which is compatible with the OpenAI Gym toolkit. 

\begin{figure}[h!]
    \centering
    \includegraphics[width=0.48\textwidth]{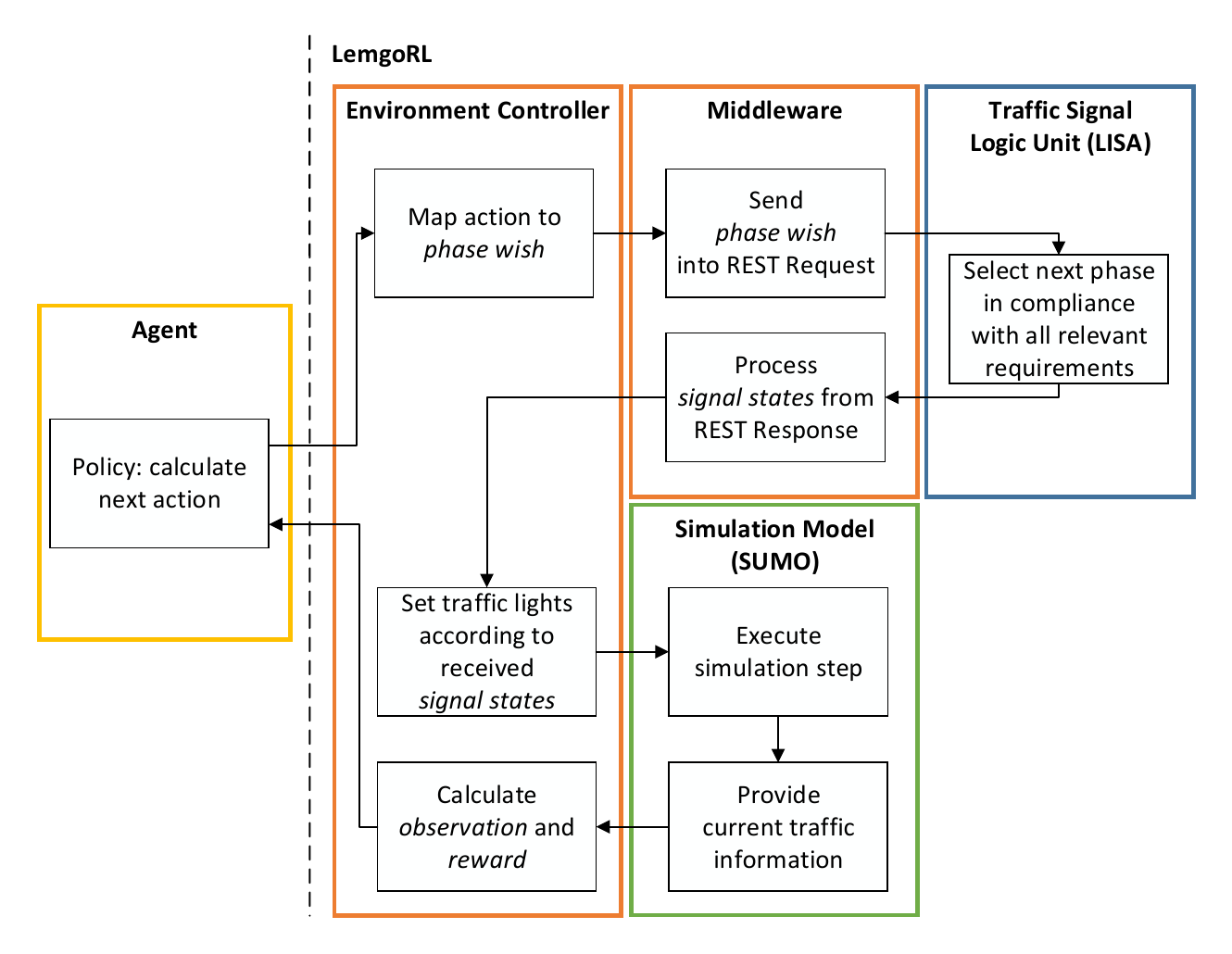}
    \caption{Process Diagram of LemgoRL}
    \label{fig:process}
\end{figure}

In Fig.~\ref{fig:process} the process diagram of LemgoRL is shown. 
An RL agent computes the next action, which represents the agent's wish for the next traffic light phase.
The environment controller passes this next action to the middleware, where it is packed into a message and sent via a REST API to the TSLU. There, it gets processed according to the logic designed by traffic engineers. Depending on the phase request and that logic, the TSLU keeps the current phase or initiates a transition to another one. Then a response message is sent back to the middleware containing the next phase with its corresponding signal states. The environment controller maps these signal states to the traffic lights in the simulation model. Following this approach, the simulated traffic lights are now in the traffic light phase, selected by the TSLU.
Subsequently, SUMO executes a step in the simulation model and provides the information of the new traffic situation to the environment controller. The latter processes the information to obtain the state and reward signal of the underlying MDP and forwards them to the agent. 

In the following subsections, the simulation model and the TSLU are explained in more detail.  

\subsection{Simulation Model}
\begin{figure}
    \centering
    \includegraphics[width=0.48\textwidth]{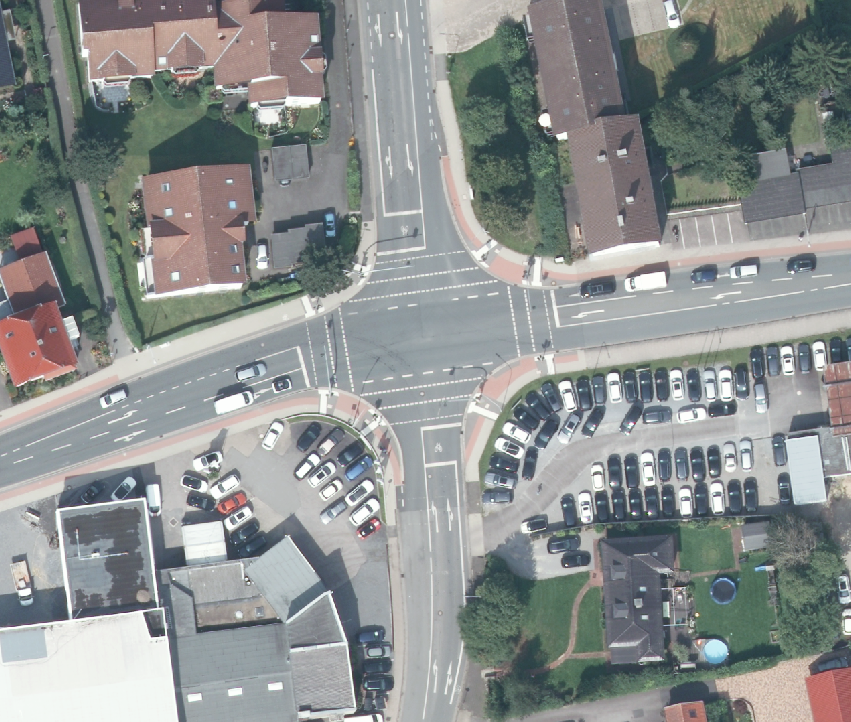}
    \caption{Satellite Photo of the intersection "OWL322" in Lemgo}
    \label{fig:owl322}
\end{figure}

\begin{figure}
    \centering
    \begin{subfigure}{0.48\textwidth}
        \centering
        \includegraphics[width=\textwidth]{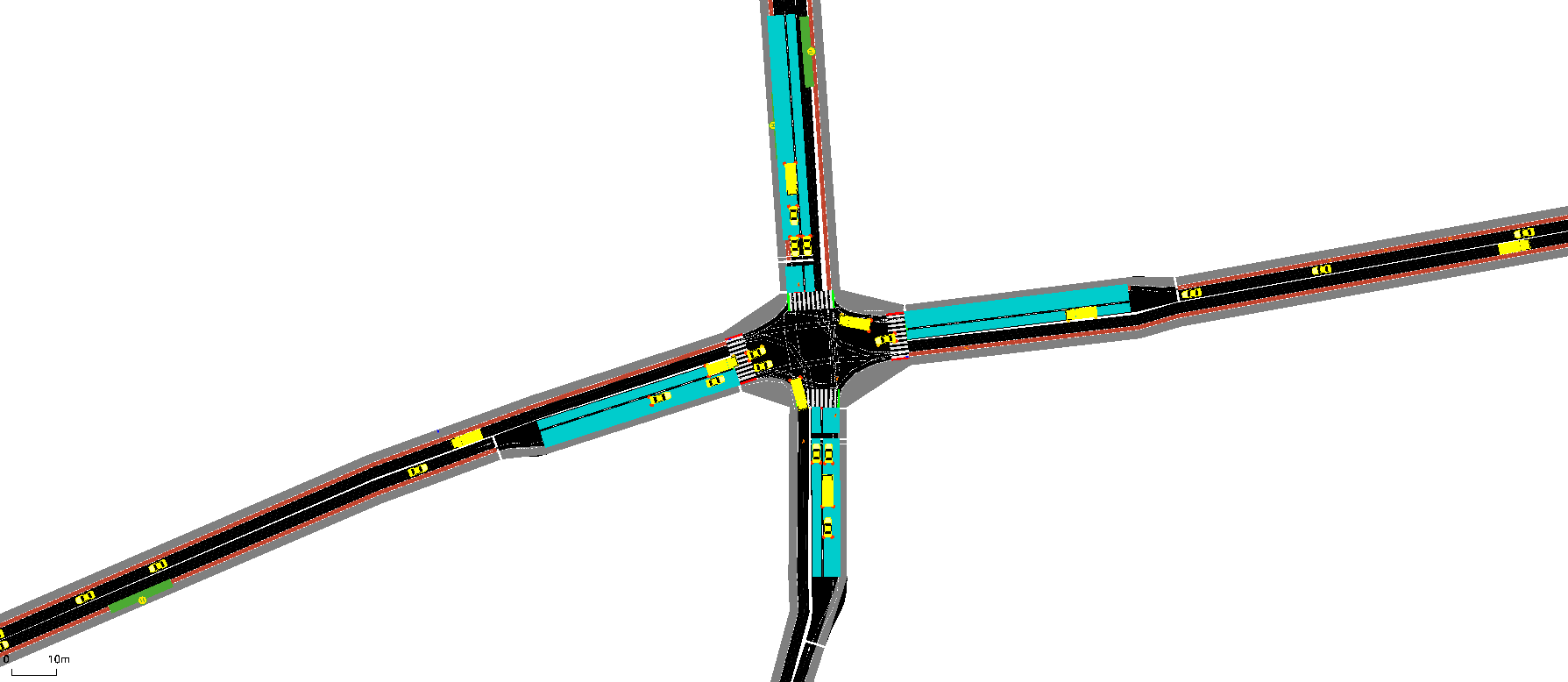}
        \caption{OWL322 distant}
        \label{fig:owl322_distant}
    \end{subfigure}
    \hfill
    \begin{subfigure}{0.48\textwidth}
        \centering
        \includegraphics[width=\textwidth]{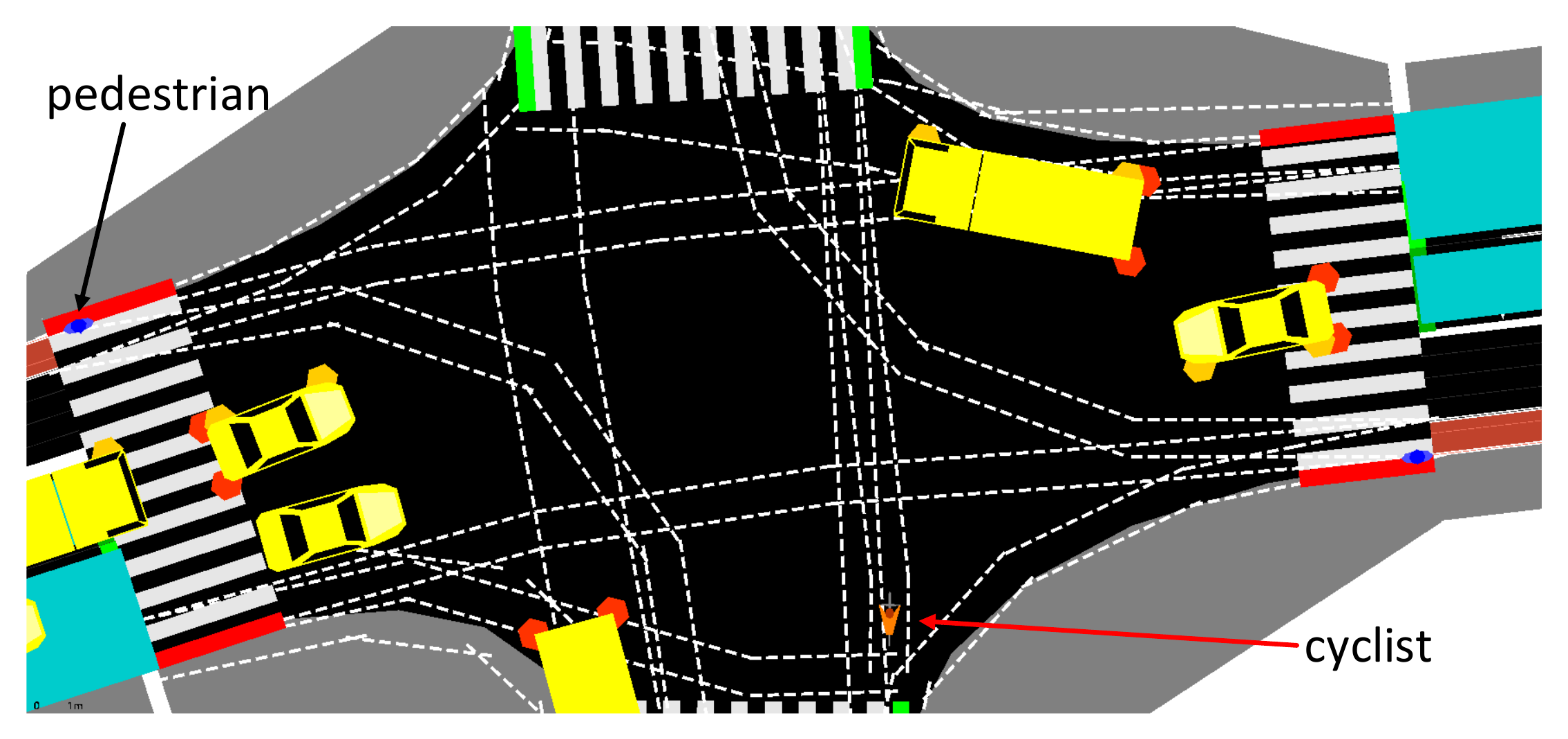}
        \caption{OWL322 close}
        \label{fig:owl322_close}
    \end{subfigure}
    \caption{OWL322 intersection SUMO model}
    \label{fig:owl322_model}
\end{figure}

We built a simulation model for the intersection "OWL322" and its immediate vicinity in Lemgo (see Fig.~\ref{fig:owl322}). The intersection is known to suffer from congestion. 
The majority of traffic flows in the west-east direction. The traffic volume to and from the south is relatively low. All four incoming streets each have a shared lane for straight and right-turning traffic, as well as a separate left-turn lane. On the west and east street, bicycles share the road with vehicles. On the north and south street, bicycles have a separate lane with a dedicated waiting area in front of the vehicles.
Furthermore, the intersection includes pedestrian crossings in all four directions and bus stations nearby, which are part of the simulation model as well. The simulation model is shown in Fig.~\ref{fig:owl322_model}. It consists of several files encoding information for the road network data, traffic demand, and traffic infrastructure, referred to as the \textit{simulation scenario} in the SUMO setting \cite{Lopez2018}. 

The default way of communicating with a running SUMO instance is TraCI, which is based on a TCP client/server architecture. However, there is an alternative called Libsumo\footnote{https://sumo.dlr.de/docs/Libsumo.html}, which we also integrated into LemgoRL. It is a reimplementation of TraCI with static C++ functions avoiding the need for socket communication. Based on our observations, this brings an increase in performance of up to a factor of 10. 
However, the usage of Libsumo requires the additional effort of compiling SUMO from source. This is why we make it optional so that a user can choose between TraCI and Libsumo.

%

\subsection{Traffic Signal Logic Unit}    
In order to have a realistic simulation environment, mechanisms are necessary to guarantee compliance with all safety and regulatory requirements. To this end, we developed a TSLU incorporating these mechanisms. The logic for these mechanisms was implemented by traffic engineers from our team, knowing and being able to apply all relevant requirements. The TSLU is based on the professional
traffic engineering software LISA (Version 7.2)\footnote{https://www.schlothauer.de/en/software-systems/lisa/}, so that the engineers were able to implement the logic in a domain-specific tool. LISA is a widely used and established commercial software package for planning, developing, and evaluating traffic signal controllers, that can be deployed in real-world traffic light systems.

\section{Features of LemgoRL}
\label{sec:features}
In the following section, we want to highlight some features of LemgoRL.




\subsection{Realistic Traffic Demand}
\begin{figure}[ht!]
    \centering
    \includegraphics[width=0.48\textwidth]{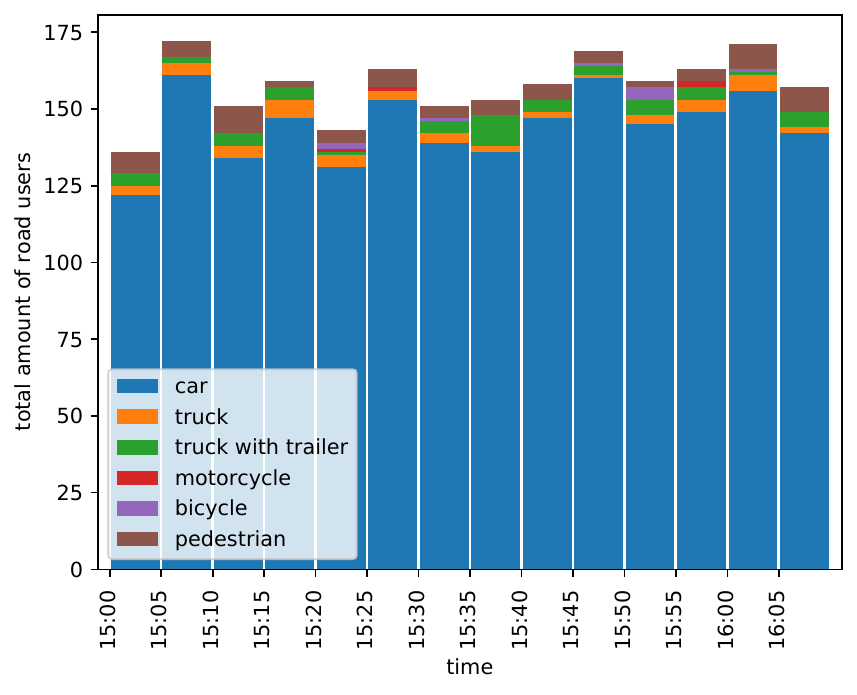}
    \caption{Measured Traffic Volume}
    \label{fig:traffic_volume}
\end{figure}

In order to increase the realism in our simulation scenario we measured the actual traffic in the OWL322 intersection during the afternoon rush hour (from 15:00 to 16:10) and transferred the results to the model. We counted the number of cars, motorcyclists, trucks, trucks with trailers, buses, and bicyclists in 5-minute periods according to their respective directions of travel. The usage of these vehicle categories for measuring the traffic volume is recommended by the German Federal Ministry of Transport \cite{BAST}. We also counted the number of pedestrians crossing the street in the same time periods. The total number of all road users over time is shown in Fig.~\ref{fig:traffic_volume}.

From the measured traffic data, we calculated the "spawning" probabilities for all vehicle types and pedestrians per second and integrated them in the model as SUMO "flows". The underlying probability distribution for a vehicle or pedestrian to spawn is the binomial distribution. By using these probabilities, each training iteration exhibits a slightly different traffic demand and dynamics. This provides the RL agent a richer dataset during training, potentially leading to more robust policies.
We have open-sourced the measured data as well as a Python script to convert this data to SUMO flows along with LemgoRL. 
To further increase the realism of the model, we added all bus stops near the intersection and buses driving according to the routes and schedule of the local transport companies\footnote{https://teutoowl.de/owlv/, https://www.stadtbus-Lemgo.de/}.

\subsection{MDP Specification}
LemgoRL formalizes the problem of controlling the OWL322 traffic lights as an MDP in order to be solvable with RL algorithms. The MDP is episodic with one episode lasting $\SI{4200}{\second}$, which corresponds to the recorded time interval of the traffic measurement. Unlike most other related work, we select a short interaction time of $\SI{1}{\second}$ for interactions between the agent and its environment. This interaction time resembles the common processing time of traffic light systems in Germany, where the entire cycle from processing the detector values to setting the traffic light phases takes $\SI{1}{\second}$.

\subsubsection{State Definition}
We define the state as
\begin{multline}
    s_{t} = \{queue_{t,l}, wave_{t,l}, avg\_speed_{t,l}, wait\_veh_{t,l},\\
    wait\_ped_{t,l}, cur\_phase\_id_{t}, des\_phase\_id_{t}, \\ 
    time\_elapsed\_curphase_{t}, time\_elapsed\_desphase_{t},\}_{l\in L}
\end{multline} 
where $l$ denotes an incoming lane or pedestrian crosswalk of the OWL322. 
The definitions for $wave$ and $wait\_veh$ are taken from \cite{Chu2019}. $wave$ is the number of nearing vehicles per lane, whereas $wait\_veh$ represents the waiting time of a lane's leading vehicle. $queue$ measures the length of all waiting vehicles in a lane (queue length), while $avg\_speed$ is the average speed of all vehicles in the respective lane.
The waiting time for the first person waiting for a green signal at a crosswalk is given by $wait\_ped$. Furthermore, we also include a one-hot encoded vector representing the current phase $cur\_phase\_id$ and one representing the phase desired by the agent $des\_phase\_id$. Additionally, we include the elapsed time in the current phase $time\_elapsed\_curphase_{t}$ and how long the agent has been desiring its phase $time\_elapsed\_desphase_{t}$.

We provide this rich state representation for two reasons. First, the proper state space design for TSC is still an open research question \cite{Haydari2020}. By providing this variety, researchers can select a subset as state space representations tailored to the requirements of the used algorithms. Second, the analysis conducted in \cite{Alegre2021} indicates that a broader scope of information in the state signal leads to better performance when facing non-stationarity due to changes in traffic dynamics. The latter is the case in real-world scenarios like LemgoRL.  


\subsubsection{Action Definition}
Modern traffic light systems are usually controlled by a phase-oriented approach, i.e. the controller selects traffic light phases based on the current traffic situation or according to a fixed time scheme. A traffic light phase represents a combination of signal states (i.e. signal colors like red or green). It determines which non- or partially conflicting traffic flows are allowed to cross the intersection and which have to wait \cite{FGSV2015}. One phase for example would allow traffic on the west-east direction while stopping traffic on the north-south direction. 

In LemgoRL, traffic engineers designed a set of 8 traffic light phases that are used in the current real-world implementation of the OWL322 (see table \ref{tab:phases}). 
In phase 3, vehicles are allowed to travel on the west-east axis, while in phase 2, pedestrians are additionally allowed to cross the street in the north and south. 
Equivalently, north-south traffic is allowed for vehicles in phase 6, with pedestrians additionally allowed to cross the streets in the west and east in phase 7.
Phase 4 and 8 only allow traffic coming from the west and north, respectively, reflecting the high traffic demand from these directions.
Noteworthy is also phase~5 with a priority green signal for left turning traffic coming from west. It can only be reached when transitioning from phase~4 after waiting for several seconds to allow traffic (which came from the east) to clear the intersection. This phase handles the large traffic demand coming from the west and heading north. 
Phase~1 represents the "all-red-phase". It is required as an intermediate phase to ensure safe transitions, e.g. when transitioning from phase~3 (vehicles west-east) to phase~7 (vehicles north-south). Except for phase~1, all phases are accessible to the agent, so the action space comprises 7 discrete actions.
\begin{table}
\renewcommand{\arraystretch}{1.3}
\centering
\begin{threeparttable}
\caption{Traffic Light Phases for OWL322}
\label{tab:phases}
 \begin{tabular}{|c || c | c | c | c| c | c | } 
 \hline
 \bfseries phase & \bfseries veh & \bfseries veh & \bfseries veh & \bfseries veh & \bfseries ped & \bfseries ped \\ 
 \bfseries nr & \bfseries west & \bfseries east & \bfseries north & \bfseries south & \bfseries w\&e & \bfseries n\&s \\ 
 \hline\hline
 1 & r & r & r & r & r & r \\ 
 \hline 
 2 & g & g & r & r & r & g \\ 
 \hline
 3 & g & g & r & r & r & r \\ 
 \hline
 4 & g & r & r & r & r & r \\ 
 \hline
 5 & g+left~g & r & r & r & r & r \\ 
 \hline
 6 & r & r & g & g & g & r \\ 
 \hline
 7 & r & r & g & g & r & r \\ 
 \hline
 8 & r & r & g & r & r & r \\ 
 \hline
\end{tabular}
\begin{tablenotes}
\item [*]vehicle traffic signal encompasses all directions (e.g. veh~west=all vehicles coming from west and traveling to the north, east or south) 
\item [**]r=red, g=green, left~g=left turning vehicles have priority green 
\end{tablenotes}
\end{threeparttable}
\end{table}


\subsubsection{Reward Function}
For the reward, we extend the definition in \cite{Chu2019} by including the waiting time for pedestrians $wait\_{ped}$: 
\begin{multline}
    r_{t+1} = - \sum_{l\in L} (queue_{t+1,l}  +  \alpha_{veh}\cdot wait\_veh_{t+1,l} \\ + \alpha_{ped}\cdot wait\_ped_{t+1,l})
\end{multline} 
By adjusting the tradeoff coefficients $\alpha_{veh}$ and $\alpha_{ped}$, the agent can be influenced to prioritize pedestrians over vehicles or vice versa. This could be used, for example, in congested cities to provide incentives for people not to use cars.  


\subsection{Realistic Traffic Light Phase Transitions}

Whenever the agent selects a phase other than the current one, the TSLU checks whether the transition is allowed or not. For example, if the current phase is 2 (vehicles west-east, pedestrians north and south), a phase transition is only permitted if the minimum green time for pedestrians crossing the street is not undercut. Otherwise, a phase transition would violate safety requirements, as pedestrians would not have sufficient time to safely cross the street. In this case, the TSLU would ignore the agent's phase wish and would stay in the current phase.

Designing proper phase transitions is a sophisticated traffic engineering task. It requires determining all transition and intergreen times for "all combinations of conflicting traffic flows"~\cite{FGSV2015}. The intergreen times are composed of crossing, clearance and entering times. They in turn depend on the permissible speed, the road user groups and actual topology and geometrical dimensions of the intersection. In contrast to the related work, where simplified phase transitions are used, LemgoRL provides realistic phase transitions for all possible combinations. 

\subsection{Transferability to Real-world Application}
\begin{figure}
    \centering
    \includegraphics[width=0.48\textwidth]{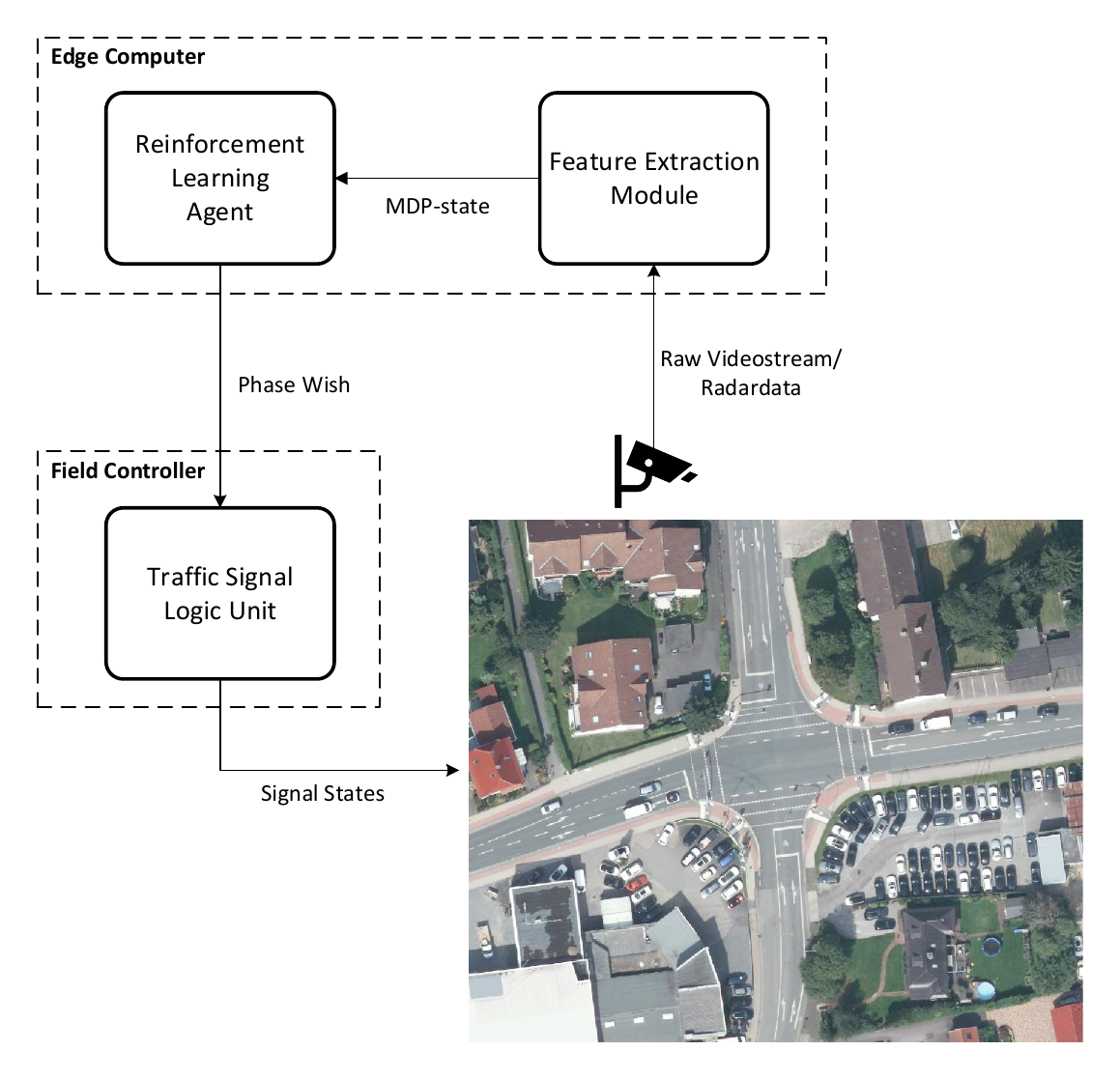}
    \caption{Schema for Real-World Deployment of LemgoRL}
    \label{fig:deployment}
\end{figure}

The ultimate goal of our research work is to deploy an RL agent trained with our benchmark tool as real-world TSC at the OWL322 intersection. Therefore, we designed our benchmark tool in way that supports this goal. By providing the agent with the real traffic phases, ensuring proper phase transitions by the TSLU and ensuring compliance to all safety and regulatory requirements at all times, the trained agent in combination with the TSLU can be deployed at the real OWL322 intersection. In Fig.~\ref{fig:deployment}, the schema for a real-world deployment is shown. It should be noted, that the current sensors at the OWL322 intersection are not sufficient to capture all necessary information for calculating the MDP state signal. Therefore, additional camera or radar sensors need to be installed. 

The workflow of the real-world deployment is similar to that of LemgoRL. The raw data of the sensors will be processed in a feature extraction module running on an edge computer. It converts the data to MDP state signals. The state signal must be identical in dimensions with that from the training process in order to be applicable to the RL agent. Based on this state signal, the RL agent selects the next phase which will be handed over to the TSLU running on the field controller of the traffic light system. Finally, the TSLU sets the state signals for all traffic light devices of the intersection.

\section{Evaluation}
To demonstrate the functionality of LemgoRL we train an RL agent with our benchmark tool and compare its performance with baseline methods. 

\subsection{Experimental Setup}

The traffic simulation settings follow the descriptions in Section~\ref{sec:features}. To extract all information needed for the state and reward of the MDP, we use SUMO`s \textit{laneAreaDetectors} on each incoming lane, which resemble camera sensors in real-world applications. The length of this detectors is $\SI{45}{\meter}$ in the west, east and north. In the south its length is $\SI{30}{\meter}$, since the road has a bend here. However, due to low traffic volume from the south, the laneAreaDetector is still able to track all incoming vehicles so that the state and reward do not get distorted.  

We evaluate our RL agent and baseline policies over 20 test episodes. Each episode is generated with a different random seed, but the same random seeds are used for each policy to ensure a fair comparison. All methods are evaluated by the performance metrics average queue length, average waiting time for vehicles, and average waiting time for pedestrians, as these are the components of the reward function. Additionally, we add the average cumulative reward since this represents the overall performance of all three aforementioned metrics and the average speed. 

\subsection{RL and Baseline Policies}
We use Proximal Policy Optimization (PPO) \cite{schulman2017}, which is a state-of-the-art deep reinforcement learning algorithm, to show the functionality of our benchmark tool. Since the performance of deep RL algorithms is sensitive to their hyperparameters, as shown in \cite{Genders2019}, we conduct a random search with 20 trials. Every agent is trained with $2000$ episodes. The training is performed on a CPU cluster with 32 cores utilizing the RLlib framework \cite{Liang2018} for distributed and parallel training on the cluster. As a result, the training time could be reduced drastically. Training the agent on a single core takes around $\SI{780}{\minute}$, while training on all 32 cores only around $\SI{90}{\minute}$. 
The hyperparameters of the best performing agent are shown in table~\ref{tab:hyperparameters}. For clarity reasons, the nomenclature follows that of RLlib. Another important aspect for successfully training an agent is normalization. Therefore we normalize the components in the state and reward function as shown in table~\ref{tab:hyperparameters}. After normalization, all values lie in the same range.

\begin{table}[h!]
\centering
\caption{Hyperparameters used for PPO algorithm and MDP}
\label{tab:hyperparameters}     
\begin{tabular}{l l}
\hline
\textbf{Hyperparameter} & \textbf{Value} \\
\hline
train\_batch\_size & $8000$ \\
num\_sgd\_iter & $10$ \\
gamma & $0.98$ \\
lambda & $0.95$ \\
vf\_loss\_coeff & $0.1789$ \\
lr & $1.5e-5$ \\
\hline
queue\_norm & $30$ \\
wave\_norm & $14$ \\
speed\_norm & $14$ \\
wait\_veh\_norm & $14$ \\
wait\_ped\_norm & $10$\\
elapsed\_norm & $10$\\
$\alpha_{veh}$ & $1$ \\
$\alpha_{ped}$ & $0.25$ \\
\hline
\end{tabular}
\end{table}



We compare the RL agent with three rule-based baselines to better assess its performance.
\begin{itemize}
    \item \textit{Fixed Timings Policy}: The traffic signals are switched in a fixed sequence, with the phase durations corresponding to that of the current control program for afternoon rush hour. 
    \item \textit{Random Timings Policy}: The traffic signals are switched in the same fixed sequence as in the Fixed Timings Policy, but the phase durations are sampled uniformly at random.
    \item \textit{Greedy/Longest-queue-first Policy}: At each timestep, the traffic signal is selected that minimizes the immediate queue length.
\end{itemize}

\subsection{Experimental Results}
\begin{figure}[ht!]
    \centering
    \includegraphics[width=0.495\textwidth]{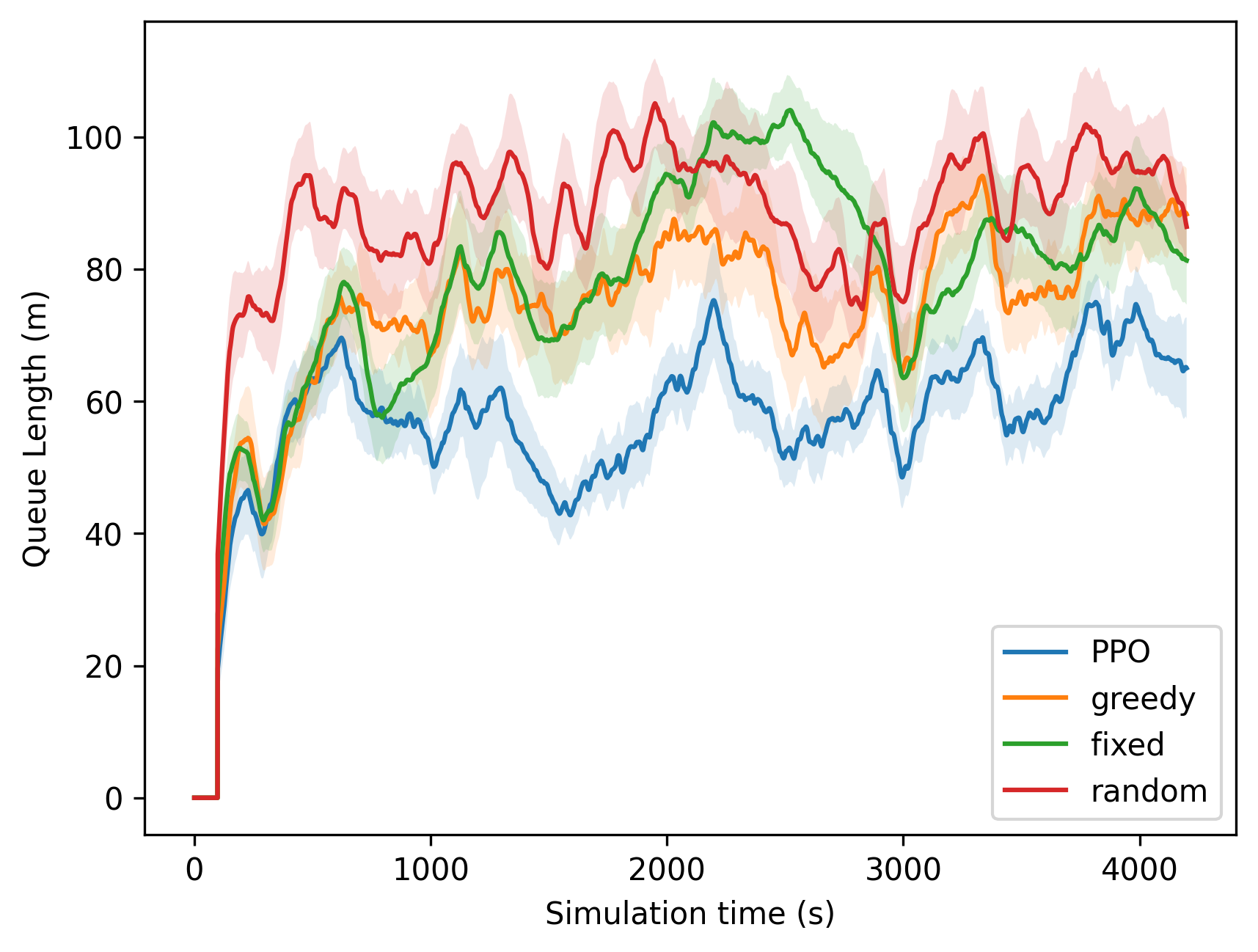}
    \caption{Queue Length\textsuperscript{5}}
    \label{fig:results_ql}
\end{figure}

\begin{figure}[ht!]
    \centering
    \includegraphics[width=0.495\textwidth]{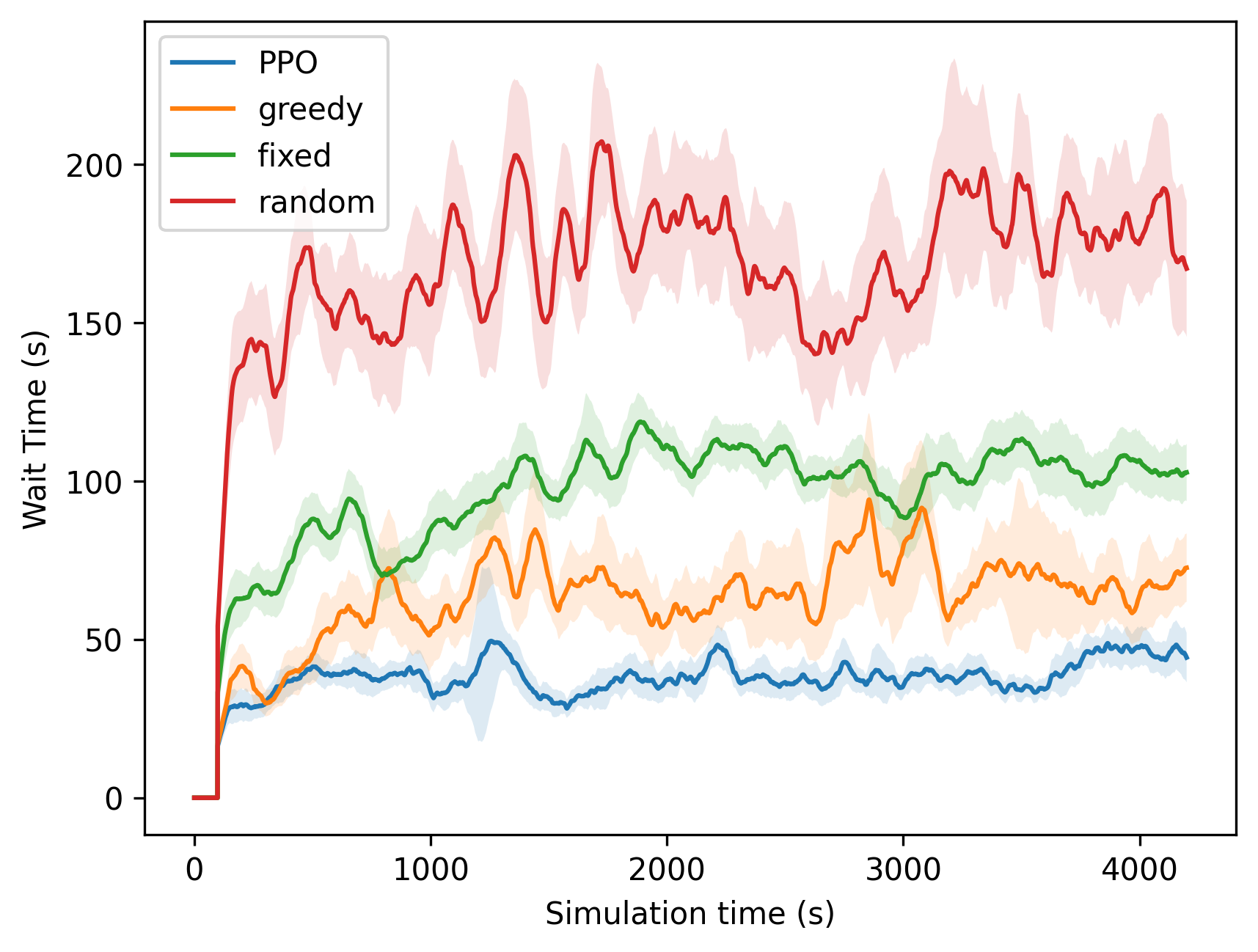}
    \caption{Waiting Time for Vehicles\textsuperscript{5}}
    \label{fig:results_veh_wait}
\end{figure}

Fig.~\ref{fig:results_ql} and fig.~\ref{fig:results_veh_wait} plot the vehicle related performance metrics, queue length and waiting time for vehicles, over the simulation time for all policies. The RL agent outperforms the baseline policies in both metrics. E.g., the average waiting time for vehicles is $\SI{37.9}{\second}$ for PPO (see table~\ref{tab:results}). Compared to greedy policy, which performs second best with $\SI{63.1}{\second}$, this is a reduction by $39.9\%$.

\begin{table}[h!]
\centering
\caption{Performance Comparison. Best values are in bold.}
\label{tab:results}     
\begin{tabular}{l|c c c c c}
\hline
Metrics & Random & Fixed & Greedy & PPO \\
\hline
cumulative reward & -5834.7 & -3756.5 & -3295.5 & \textbf{-2269.7}  \\
avg. queue length [m] & 88.2 & 78.5 & 74.5 & \textbf{57.9} \\
avg. waiting time veh. [s] & 166.7 & 96.1 & 63.1 & \textbf{37.9}  \\
avg. waiting time ped. [s] & 124.6 & 16.2 & 8.2 & \textbf{2.0}  \\
avg. vehicle speed [m/s] & 7.5 & \textbf{8.5} & 7.3 & 7.8  \\
\hline
\end{tabular}
\end{table}

\begin{figure}[ht!]
    \centering
    \includegraphics[width=0.495\textwidth]{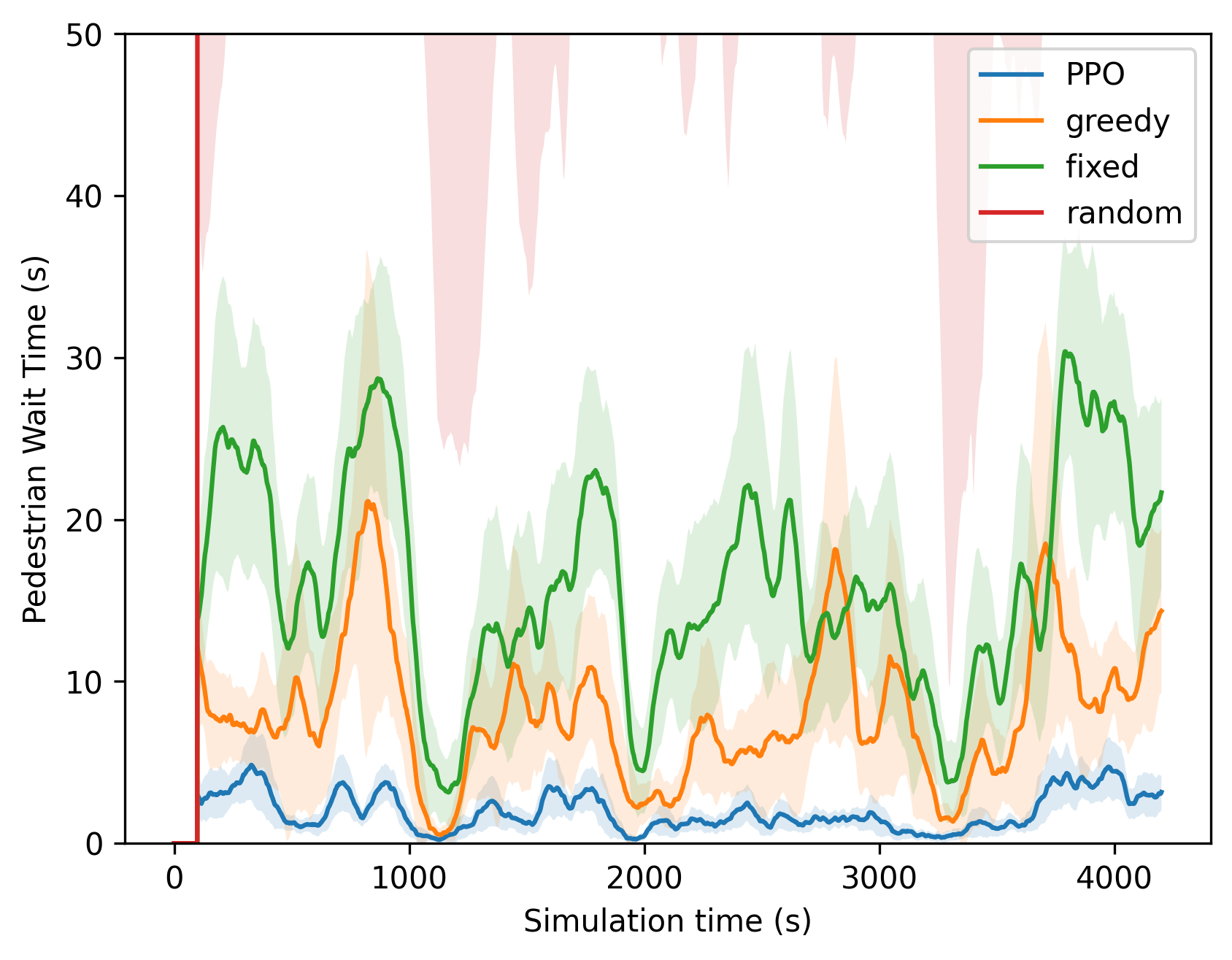}
    \caption{Waiting Time for Pedestrians\textsuperscript{5}} 
    \label{fig:results_ped_wait}
\end{figure}

When it comes to average waiting time for pedestrians, as shown in fig.~\ref{fig:results_ped_wait}, the RL agent significantly beats the other methods. In comparison to the greedy policy, PPO reduces the average waiting time for pedestrians by $75.6\%$ and compared to fixed timings policy even by $87.7\%$. The main reason for this enormous difference is probable that the RL agent incorporates the minimization of the waiting time for pedestrians as optimization goal, while the rule-based policies focus only on vehicles. It is also noteworthy, that PPO performs better in both, vehicle and pedestrian related metrics, although these are opposing each others. This highlights the optimization potential that RL holds for TSC.

\footnotetext[5]{Shaded area represents standard deviation. To smooth-out the curves, we use a moving average with the window size of 100.}

It is interesting to notice that the average vehicle speed is close to each other for all policies, even for the random timing policies, which performs significantly worse when it comes to average waiting time for vehicles and pedestrians. Fixed timings policy achieves the best results for average vehicle speed, performing $9\%$ better than the RL agent.




\section{Conclusion}
In this paper, we have presented LemgoRL, an open-source benchmark tool to train RL agents as TSC in a realistic simulation environment. The experimental results demonstrate, that a state-of-the-art deep reinforcement learning algorithm like the PPO algorithm can be trained with our benchmark tool and outperform rule-based methods. Furthermore, LemgoRL was shown to work with a modern, high-performance RL framework such as RLlib, enabling parallelization so that e.g. training time can be drastically reduced. 

By enabling researchers to train and evaluate algorithms in such a realistic environment and by providing a blueprint of how to ensure compliance with all regulatory and safety requirements, this tool drives the development of RL algorithms for traffic signal control towards real-world deployment. In future work, we plan to deploy agents trained with our benchmark tool on the real-world OWL322 intersection. In this way, we aim to prove the applicability and superiority of RL-based TSC in a real-world deployment.






\section*{Acknowledgment}
The authors would like to thank Mirko Barthauer\footnotemark[6],
Jakob Erdmann\footnotemark[7], Maximiliano Bottazi\footnotemark[7],
and the company Schlothauer \& Wauer for giving us helpful information about the interface to the LISA controller.

\bibliographystyle{IEEEtran.bst} 
\balance
\bibliography{KI4LSA.bib, motivation.bib, simulators_sumo.bib, paper1.bib, marl_and_mdp.bib, sim2real.bib}

\footnotetext[6]{Technical University of Braunschweig}
\footnotetext[7]{German Aerospace Center}

\end{document}